\documentclass{article}
\usepackage{spconf,amsmath,graphicx,hyperref}
\usepackage{booktabs}
\usepackage{multirow}
\usepackage{subcaption}
\usepackage{graphicx}
\usepackage{romannum}

\title{When Wider Views Fail: Stress-Testing Feed-Forward 3D Reconstruction}
\name{Daisy Li$^{1}$, Kyle Gao$^{2}$, Quanyun Wu$^{1}$, Hanna Chomko$^{1}$, John S. Zelek$^{1}$, Jonathan Li$^{1}$}
\address{$^{1}$University of Waterloo \quad
$^{2}$Aalto University}

\begin{document}
%
\maketitle
\begin{abstract}
Feed-forward 3D reconstruction models enable efficient geometry estimation from sparse images, but their pretrained nature can make them vulnerable to distribution shifts beyond their training data. Identifying these failure modes is important for understanding when such models can be reliably deployed in unconstrained imaging settings. We investigate viewpoint variation as a controlled distribution shift by varying the angular span of sparse image inputs while keeping the input budget fixed. Across multiple feed-forward reconstruction models, we observe substantial degradation as viewpoint span increases, with wide spans producing both incomplete surface coverage and geometry unsupported by the observed imagery. These results reveal that viewpoint variation can induce failure modes beyond conventional reconstruction incompleteness, highlighting the need to evaluate pretrained feed-forward models under distribution shifts that challenge their learned geometric priors.

\end{abstract}

\section{Introduction}
\label{sec:introduction}

Reconstructing three-dimensional scenes from sparse images requires accurate local geometry and consistent alignment across views~\cite{sfm,dust3r}. Recent feed-forward approaches, including VGGT~\cite{vggt}, Pi3X~\cite{pi3,pi3x}, and VGGT-$\Omega$~\cite{vggtomega}, efficiently estimate scene geometry and camera parameters from unposed images. For urban scene reconstruction, however, their reliability also depends on handling substantial viewpoint variation as cameras move around the target structure.

Orbital capture presents a trade-off between surface coverage and inter-view consistency. Wider coverage reveals additional facades, but, under a fixed image budget, also increases the spacing between observations. Changes in visibility, appearance, and occlusion can then reduce cross-view overlap and complicate correspondence~\cite{mast3r, multisession}, potentially producing distorted geometry or spatially separated copies of the same building~\cite{repeated}. Such failures require evaluating not only surface coverage, but also whether the reconstructed geometry forms a coherent scene.

We study this behavior by evaluating VGGT, Pi3X, and VGGT-$\Omega$ on two scenes using ten-image trajectories spanning $30^\circ$ to $360^\circ$. Each complete prediction, including duplicated components, is aligned to a pseudo-reference with a single similarity transformation and evaluated using distance- and coverage-based metrics. VGGT and Pi3X degrade substantially as the span increases, while VGGT-$\Omega$ remains robust, preserving both surface completeness and global scene coherence, with markedly stronger agreement with the pseudo-reference. These results identify wide-span orbital capture as an important stress test for the global coherence of feed-forward reconstruction.

\section{Related Work}
\label{sec:related_work}

Traditional structure-from-motion and multi-view stereo pipelines combine feature matching, camera estimation, triangulation, and global optimization. Recent learning-based methods instead regress scene geometry directly from unposed images. DUSt3R~\cite{dust3r} introduced pairwise point-map regression followed by global alignment, while MASt3R~\cite{mast3r} augmented this representation with explicit dense matching. Fast3R~\cite{fast3r} extended point-map prediction to many images in a single forward pass, and CUT3R~\cite{cut3r} introduced a persistent state for online reconstruction. More recent unified models jointly estimate multiple geometric quantities: VGGT~\cite{vggt} predicts cameras, depth, point maps, and tracks; Pi3~\cite{pi3}, subsequently extended as Pi3X~\cite{pi3x}, adopts permutation-equivariant multi-view reasoning; and VGGT-$\Omega$~\cite{vggtomega} improves cross-view reasoning through register-based information exchange and substantially expanded training.

Recent work has also examined reconstruction under challenging viewpoint changes. Zhang et al.~\cite{extreme} study extreme-view geometry primarily through relative pose estimation between image pairs with little or no visual overlap. Related approaches such as Glob3R~\cite{glob3r} address drift and inconsistency through explicit global optimization. In contrast, we evaluate the native feed-forward behavior of multi-view orbital reconstruction, characterizing how increasing angular coverage under a fixed image budget affects global scene coherence.

\begin{figure*}[t]
    \centering
    \includegraphics[width=\textwidth]
        {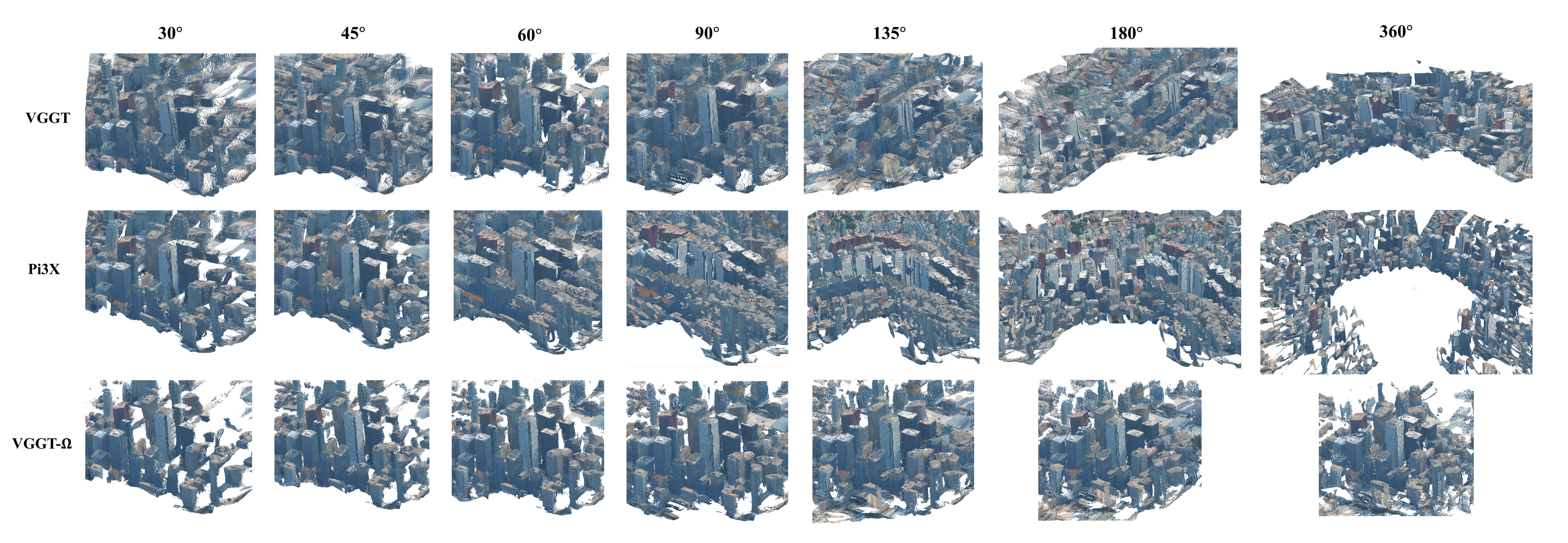}
    \caption{
    Visual comparison of Scene \Romannum{1} reconstruction results.
    }
    \label{fig:trt}
\end{figure*}
\begin{figure*}[t]
    \centering
    \includegraphics[width=\textwidth]
        {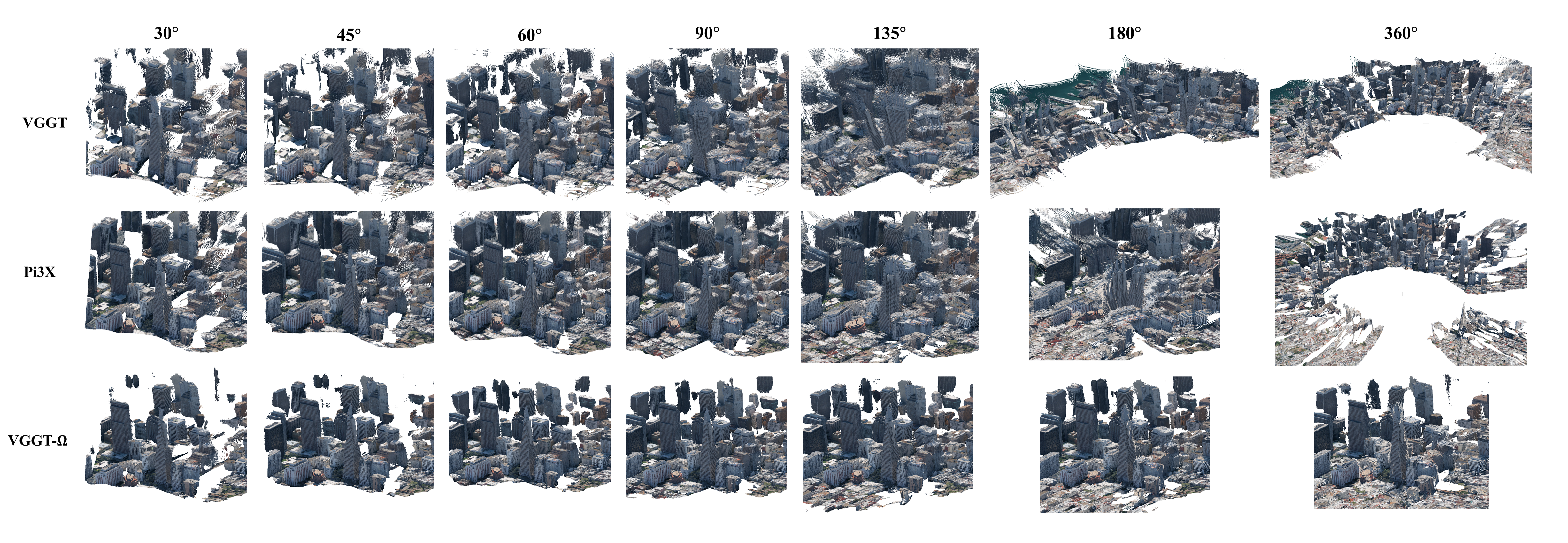}
    \caption{
    Visual comparison of Scene \Romannum{2} reconstruction results.
    }
    \label{fig:tp}
\end{figure*}

\section{Experimental Design}
\label{sec:experiment}

\subsection{Data and Viewpoint Sampling}
\label{sec:sampling}

We evaluate two urban scenes: Scene \Romannum{1}, centered on a building in downtown Toronto, and Scene \Romannum{2}, centered on the Transamerica Pyramid. The central building of Scene \Romannum{1} is characterized by predominantly vertical fa\c{c}ades, whereas the Transamerica Pyramid narrows towards its apex. This selection allows us to examine whether similar reconstruction failures occur in different urban scenes and building shapes, rather than restricting the evaluation to an isolated case.

For each scene, Google Earth Studio~\cite{googleEarthStudio} is used to generate orbital trajectories around the central building with angular spans of $30^\circ$, $45^\circ$, $60^\circ$, $90^\circ$, $135^\circ$, $180^\circ$, and $360^\circ$. Ten images are sampled uniformly from each trajectory and reconstructed independently using VGGT, Pi3X, and VGGT-$\Omega$. This produces 42 reconstruction cases in total.

\subsection{Pseudo-Reference and Alignment}
\label{sec:alignment}

For each scene, we construct a pseudo-reference by reconstructing 70 images sampled over a full $360^\circ$ orbit using VGGT-$\Omega$. The central building is then manually segmented from the scene.

The central building is also manually segmented from each ten-image reconstruction. If a model produces multiple spatially separated copies of the building, all copies are retained to ensure that reconstruction failures contribute to the evaluation. Each segmented prediction is aligned with its pseudo-reference using a single similarity transformation,
\begin{equation}
    \mathcal{T}(\mathbf{x})
    = s\mathbf{R}\mathbf{x}+\mathbf{t},
    \qquad \mathbf{R}\in SO(3),\;s>0,
    \label{eq:sim3}
\end{equation}
estimated through coarse registration followed by trimmed ICP refinement~\cite{method,trimmed}. The same transformation is applied to the entire prediction; duplicated components are not aligned independently or removed.

\begin{table}[t]
\centering
\caption{Quantitative results of Scene~\Romannum{1} across all angular spans.
CD and EMD denote normalized distances multiplied by $10^3$.
Precision (P), recall (R), and F-score (F) are percentages.
Bold F-scores indicate the best model at each span.}
\label{tab:toronto_all}
\small
\setlength{\tabcolsep}{4pt}
\begin{tabular}{llrrrrr}
\toprule
Model & $\Theta(^\circ)$
& CD$\downarrow$ & EMD$\downarrow$
& P$\uparrow$ & R$\uparrow$ & F$\uparrow$ \\
\midrule

\multirow{7}{*}{VGGT~\cite{vggt}}
& 30  & 8.63   & 66.12   & 93.34 & 66.21 & 77.47 \\
& 45  & 7.80   & 60.95   & 94.00 & 73.18 & 82.30 \\
& 60  & 13.39  & 101.28  & 80.72 & 64.18 & 71.51 \\
& 90  & 21.97  & 161.16  & 70.63 & 61.80 & 65.92 \\
& 135 & 15.81  & 109.02  & 47.28 & 76.32 & 58.39 \\
& 180 & 356.42 & 794.98  & 6.12  & 27.16 & 9.99 \\
& 360 & 534.33 & 1104.38 & 15.59 & 49.36 & 23.69 \\
\midrule

\multirow{7}{*}{Pi3X~\cite{pi3,pi3x}}
& 30  & 7.32    & 64.18   & 96.43 & 76.95 & \textbf{85.59} \\
& 45  & 7.27    & 55.61   & 90.01 & 79.47 & \textbf{84.41} \\
& 60  & 11.46   & 64.51   & 71.30 & 74.47 & 72.85 \\
& 90  & 38.07   & 117.67  & 44.22 & 65.69 & 52.86 \\
& 135 & 173.48  & 413.36  & 7.94  & 45.83 & 13.54 \\
& 180 & 433.85  & 899.56  & 9.11  & 43.46 & 15.06 \\
& 360 & 1190.54 & 2356.08 & 8.99  & 43.02 & 14.87 \\
\midrule

\multirow{7}{*}{VGGT-$\Omega$~\cite{vggtomega}}
& 30  & 7.42  & 65.58  & 96.60 & 69.78 & 81.03 \\
& 45  & 8.84  & 77.77  & 91.84 & 77.98 & 84.35 \\
& 60  & 6.57  & 65.90  & 94.85 & 83.99 & \textbf{89.09} \\
& 90  & 8.39  & 85.80  & 86.92 & 83.25 & \textbf{85.05} \\
& 135 & 7.70  & 76.42  & 83.46 & 84.11 & \textbf{83.78} \\
& 180 & 7.29 & 61.26  & 80.05 & 84.57 & \textbf{82.25} \\
& 360 & 5.82  & 50.20  & 82.89 & 89.47 & \textbf{86.05} \\
\bottomrule
\end{tabular}
\end{table}

\begin{table}[t]
    \centering
    \caption{Quantitative results of Scene~\Romannum{2} across all angular spans.
    CD and EMD denote normalized distances multiplied by $10^3$.
    Precision (P), recall (R), and F-score (F) are percentages.
    Bold F-scores indicate the best model at each span.}
    \label{tab:tp_all}
    \small
    \setlength{\tabcolsep}{4pt}
    \begin{tabular}{llrrrrr}
        \toprule
        Model & $\Theta(^\circ)$
        & CD$\downarrow$ & EMD$\downarrow$
        & P$\uparrow$ & R$\uparrow$ & F$\uparrow$ \\
        \midrule

        \multirow{7}{*}{VGGT~\cite{vggt}}
        & 30  & 12.29  & 55.67   & 83.76 & 40.26 & 54.38 \\
        & 45  & 12.02  & 56.77   & 85.50 & 42.34 & 56.63 \\
        & 60  & 11.47  & 56.31   & 85.22 & 46.32 & 60.02 \\
        & 90  & 12.69  & 55.04   & 64.83 & 57.82 & 61.12 \\
        & 135 & 33.37  & 88.29   & 55.75 & 33.00 & 41.46 \\
        & 180 & 663.69 & 1429.48 & 18.46 & 20.81 & 19.57 \\
        & 360 & 759.92 & 1622.23 & 7.52  & 12.67 & 9.44 \\
        \midrule

        \multirow{7}{*}{Pi3X~\cite{pi3,pi3x}}
        & 30  & 7.67    & 61.54   & 99.06 & 59.88 & 74.64 \\
        & 45  & 7.15    & 60.71   & 98.23 & 61.06 & 75.31 \\
        & 60  & 6.95    & 54.62   & 93.85 & 64.33 & 76.34 \\
        & 90  & 6.01    & 49.70   & 94.72 & 71.45 & 81.46 \\
        & 135 & 9.96    & 77.30   & 61.20 & 79.53 & 69.17 \\
        & 180 & 51.20   & 127.92  & 20.68 & 36.20 & 26.33 \\
        & 360 & 1860.48 & 3849.65 & 10.01 & 32.05 & 15.26 \\
        \midrule

        \multirow{7}{*}{VGGT-$\Omega$~\cite{vggtomega}}
        & 30  & 8.18 & 73.46 & 98.94 & 63.70 & \textbf{77.50} \\
        & 45  & 8.54 & 74.10 & 96.88 & 63.43 & \textbf{76.67} \\
        & 60  & 7.28 & 53.57 & 99.30 & 68.71 & \textbf{81.22} \\
        & 90  & 7.32 & 44.31 & 99.58 & 69.56 & \textbf{81.90} \\
        & 135 & 4.40 & 38.37 & 99.10 & 81.82 & \textbf{89.63} \\
        & 180 & 4.14 & 37.64 & 96.22 & 87.08 & \textbf{91.42} \\
        & 360 & 3.82 & 36.99 & 93.78 & 95.72 & \textbf{94.74} \\
        \bottomrule
    \end{tabular}
\end{table}

\subsection{Evaluation Metrics}
\label{sec:metrics}

Let $P$ and $Q$ denote surface samples from the aligned prediction and pseudo-reference, respectively, and let $D$ be the pseudo-reference bounding-box diagonal. We report normalized symmetric Chamfer distance,
\begin{equation}
    \mathrm{CD}
    =
    \frac{1}{2D}
    \left(
    \frac{1}{|P|}\sum_{\mathbf p\in P}d(\mathbf p,Q)
    +
    \frac{1}{|Q|}\sum_{\mathbf q\in Q}d(\mathbf q,P)
    \right),
    \label{eq:cd}
\end{equation}
where $d(\mathbf{x},A)=\min_{\mathbf a\in A} \|\mathbf{x}-\mathbf a\|_2$. We additionally report normalized Earth Mover's Distance (EMD) using equal-size surface samples.

Surface precision and recall are computed using $\tau=0.01D$:
\begin{equation}
\begin{aligned}
\mathrm{P}_{\tau}
&=\frac{1}{|P|}\sum_{\mathbf p\in P}
\mathbf{1}\!\left[d(\mathbf p,Q)\leq\tau\right],\\
\mathrm{R}_{\tau}
&=\frac{1}{|Q|}\sum_{\mathbf q\in Q}
\mathbf{1}\!\left[d(\mathbf q,P)\leq\tau\right].
\end{aligned}
\label{eq:pr}
\end{equation}
Their harmonic mean gives the surface F-score. CD and EMD measure overall geometric discrepancy, whereas precision and recall distinguish unsupported predicted geometry from incomplete reference coverage.

Because the pseudo-references are generated by VGGT-$\Omega$, the reported scores measure agreement with learned references rather than absolute geometric accuracy and may contain model-dependent bias.

\section{Results}
\label{sec:results}

\subsection{Quantitative Results}
\label{sec:quantitative_results}

The quantitative results for all 42 reconstruction cases are presented in Tables~\ref{tab:toronto_all} and~\ref{tab:tp_all}. Both Scenes exhibit a consistent wide-span pattern: Pi3X and VGGT degrade substantially as the angular span increases, whereas VGGT-$\Omega$ maintains considerably stronger agreement with the pseudo-reference.

For Scene~\Romannum{1}, Pi3X performs best at $30^\circ$ and $45^\circ$, with F-scores of $85.59\%$ and $84.41\%$, respectively. Its performance then declines rapidly, reaching an F-score of $13.54\%$ at $135^\circ$. VGGT degrades more gradually at intermediate spans, but its F-score falls to $9.99\%$ at $180^\circ$. At $360^\circ$, Pi3X and VGGT obtain F-scores of only $14.87\%$ and $23.69\%$, respectively.

In contrast, VGGT-$\Omega$ remains stable across all Scene~\Romannum{1} settings. Its F-score varies within a comparatively narrow range of $81.03\%$--$89.09\%$, without the wide-span collapse observed for the other models. At $360^\circ$, it achieves its lowest normalized Chamfer distance ($5.82\times10^{-3}$) and EMD ($50.20\times10^{-3}$), together with an F-score of $86.05\%$.

Scene~\Romannum{2} exhibits the same broad separation at wide spans. At $360^\circ$, Pi3X and VGGT obtain F-scores of $15.26\%$ and $9.44\%$, whereas VGGT-$\Omega$ reaches $94.74\%$. The corresponding normalized Chamfer distances are $1860.48\times10^{-3}$, $759.92\times10^{-3}$, and $3.82\times10^{-3}$, respectively. VGGT-$\Omega$ therefore preserves strong reference agreement on both scenes, despite the different geometries of the central buildings and the different scene contexts.

\subsection{Qualitative Results}
\label{sec:qualitative_results}

Fig.~\ref{fig:trt} and Fig.~\ref{fig:tp} reveals that the wide-span degradation of VGGT and Pi3X is not limited to a gradual loss of surface detail. At large angular spans, the models fail to associate observations of the same physical building across widely separated viewpoints within a consistent global frame. Consequently, view-specific reconstructions are placed at incompatible positions or orientations, producing multiple displaced copies and severely distorted structures. The result is a visually catastrophic failure of scene assembly, even when individual facade fragments remain locally recognizable.

In contrast, VGGT-$\Omega$ consistently consolidates the input views into a single building instance across the tested spans. Although its local surface detail may vary, it avoids the exaggerated duplication and fragmentation observed in VGGT and Pi3X. This qualitative difference agrees with the substantially lower distance errors and higher surface precision of VGGT-$\Omega$ in the wide-span settings.

\section{Discussion}
\label{sec:discussion}

\subsection{Global Coherence under Wide Viewpoint Spans}
\label{sec:interpretation}
Our results suggest that wide-span orbital reconstruction challenges global scene coherence, rather than merely surface completeness. With ten input images, wider trajectories increase surface coverage but also increase inter-view spacing, potentially weakening correspondence through occlusion and appearance changes. VGGT and Pi3X produce displaced building components at large spans, whereas VGGT-$\Omega$ maintains substantially stronger pseudo-reference agreement. Because each complete prediction receives only one similarity transformation, incompatible components cannot be independently realigned to conceal these failures. The observed degradation is therefore consistent with errors in relative scene assembly, rather than only an arbitrary global coordinate frame or scale.

\subsection{Geometric Supervision and Training Exposure}
\label{sec:influence}
VGGT's separate camera, depth, and world-point heads do not architecturally enforce consistency between their predictions~\cite{vggt}. Pi3X transforms local point maps using predicted cameras, so pose errors can displace otherwise plausible surfaces; its public training implementation leaves the cross-view unprojection loss disabled~\cite{pi3x}. In contrast, VGGT-$\Omega$ couples predicted depth and cameras through an unprojection-based point loss, explicitly supervises token-level matching, and exchanges cross-view information through scene registers~\cite{vggtomega}. These mechanisms plausibly support coherent scene assembly, although our experiments do not isolate their individual effects.

Training exposure may reinforce these advantages. VGGT-$\Omega$ reports substantially expanded annotated data and self-supervised video training; its temporal-window sampling admits weakly overlapping observations~\cite{vggtomega}. This may improve robustness to sparse orbital inputs. However, outdoor imagery already appears in VGGT's and the original Pi3's training data, while Pi3X's complete training mixture remains undisclosed~\cite{vggt, pi3x}. Data diversity alone therefore cannot establish the cause of the observed performance gap.

\subsection{Limitations and Future Directions}
\label{sec:limitations}
Our study covers two Google Earth Studio scenes under a fixed input, which makes the conclusions specific and reproducible while leaving several questions open for future work. Because angular span jointly controls viewpoint spacing and visible content, it defines a useful operating axis but not a decomposition of rotation from overlap; controlled experiments that vary overlap at fixed rotation would isolate the two effects and locate scene-dependent thresholds. The pseudo-references from VGGT-$\Omega$ likewise measure agreement with a strong reference model rather than independent geometric accuracy, so evaluation against independent references and direct pose metrics is a natural next step, together with training ablations that test the proposed explanations. To address these failures, future research can use viewpoint adaptation in 3D foundation models, where the model explicitly adapts its predictions to the available viewpoint distribution and overlap. This could be combined with gradient-free iterative refinement to progressively improve geometry under challenging viewing conditions without requiring model retraining.

\section{Conclusion}
\label{sec:conclusion}
Pretrained feed-forward 3D reconstruction models can encounter distribution shifts beyond their training data, making systematic identification of their failure modes important for assessing their reliability. We studied viewpoint-dependent failure under controlled changes in orbital coverage and found substantial wide-span degradation in Pi3X and VGGT, including globally inconsistent and unsupported geometry, while VGGT-$\Omega$ showed greater robustness. These results demonstrate that large viewpoint variation can expose failure modes beyond incomplete surface coverage, highlighting viewpoint shift as an important factor when evaluating pretrained feed-forward reconstruction models. More broadly, the study motivates systematic evaluation across distribution shifts to better characterize the reliability and limitations of feed-forward 3D reconstruction.

\vfill\pagebreak

\bibliographystyle{IEEEbib}
\bibliography{refs}

\end{document}